\documentclass[journal]{IEEEtranTIE}
\usepackage{graphicx}
\usepackage{subfigure}
\usepackage{cite}
\usepackage{picinpar}
\usepackage{amsmath}
\usepackage{amsfonts}
\usepackage{url}
\usepackage{flushend}
\usepackage[latin1]{inputenc}
\usepackage{colortbl}
\usepackage{soul}
\usepackage{multicol}
\usepackage{multirow}
\usepackage{pifont}
\usepackage{color}
\usepackage{alltt}
\usepackage[hidelinks]{hyperref}
\usepackage{enumerate}
\usepackage{siunitx}
\usepackage{breakurl}
\usepackage{epstopdf}
\usepackage{pbox}

\usepackage{pifont}

\begin{document}
\title{AdaFusion: Visual-LiDAR Fusion with Adaptive Weights for Place Recognition}
\author{
	\vskip 1em
	
	Haowen Lai$^1$,
	Peng Yin$^{2,*}$,
	and Sebastian Scherer$^2$

	\thanks{
		{This work was funded by US ARL award W911QX20D0008.}
		
		{Haowen Lai is with the Department of Automation, Tsinghua University, Beijing, China. {(lhw19@mails.tsinghua.edu.cn).}
		Peng Yin and Sebastian Scherer are with Robotics Institute, Carnegie Mellon University, Pittsburgh, PA 15213, USA. {(pyin2, basti@andrew.cmu.edu)}

		Corresponding author: Peng Yin, (pyin2@andrew.cmu.edu)}.
	}
}
\maketitle
	
\begin{abstract}
    Recent years have witnessed the increasing application of place recognition in various environments, such as city roads, large buildings, and a mix of indoor and outdoor places.
    This task, however, still remains challenging due to the limitations of different sensors and the changing appearance of environments.
    Current works only consider the use of individual sensors, or simply combine different sensors, ignoring the fact that the importance of different sensors varies as the environment changes.
    In this paper, an adaptive weighting visual-LiDAR fusion method, named AdaFusion, is proposed to learn the weights for both images and point cloud features.
    Features of these two modalities are thus contributed differently according to the current environmental situation.
    The learning of weights is achieved by the attention branch of the network, which is then fused with the multi-modality feature extraction branch.
    Furthermore, to better utilize the potential relationship between images and point clouds, we design a two-stage fusion approach to combine the 2D and 3D attention.
    Our work is tested on two public datasets, and experiments show that the adaptive weights help improve recognition accuracy and system robustness to varying environments.
\end{abstract}


\definecolor{limegreen}{rgb}{0.2, 0.8, 0.2}
\definecolor{forestgreen}{rgb}{0.13, 0.55, 0.13}
\definecolor{greenhtml}{rgb}{0.0, 0.5, 0.0}

\section{Introduction}
    \IEEEPARstart{L}{ocalization} plays a significant role in much of the robotic applications such as autonomous driving, package delivery and service robots. Place recognition, as a typical localization method that retrieves the most similar places regarding the query one, is thus extensively studied by researchers. There are some challenges in this task, one of which comes from the potential application in various environments, making it hard to be coped with. The capability to distinguish and select prominent information thus becomes a key way to improve the recognition accuracy.
    
    \begin{figure}[!t]
    	\centering
        \includegraphics[width=\linewidth]{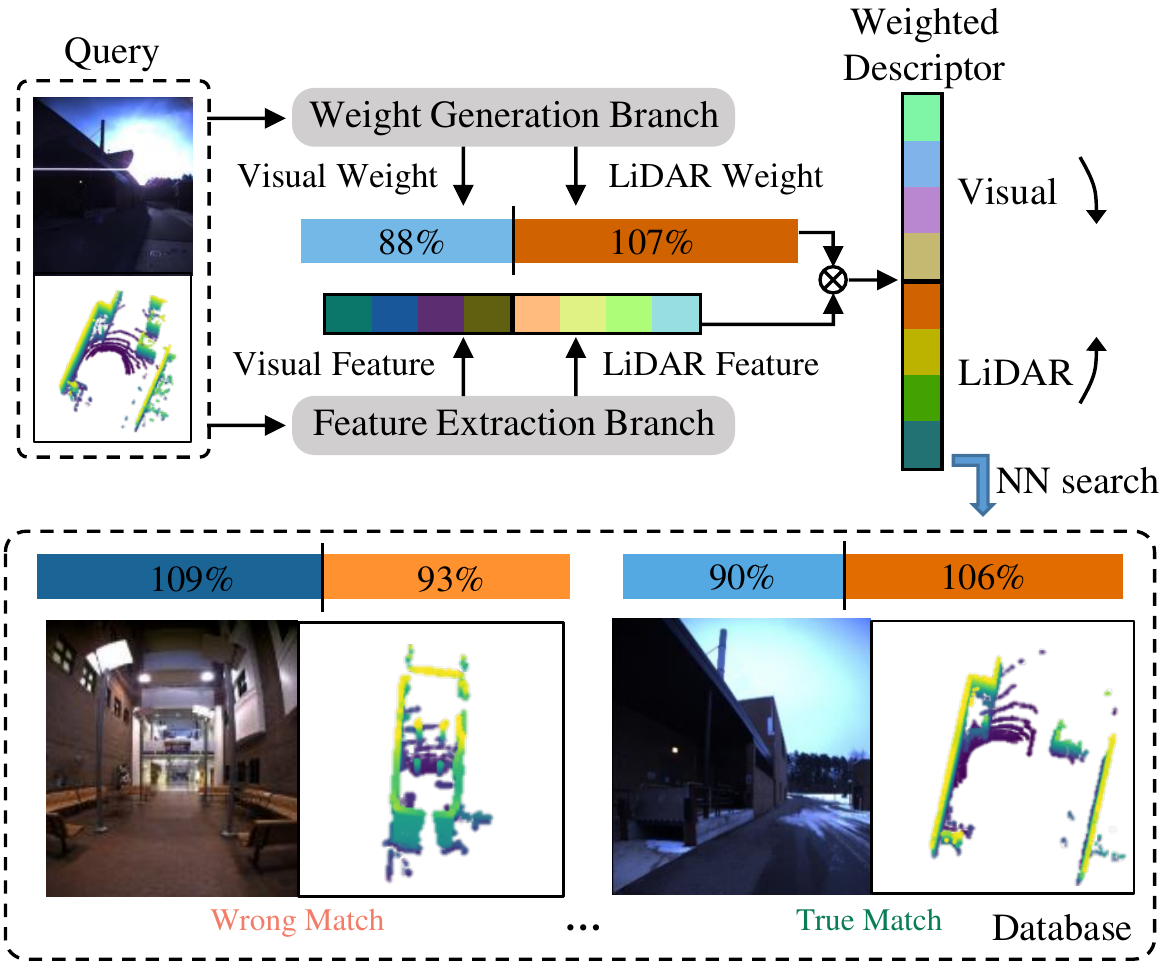}
    	\caption{\textbf{Place recognition with adaptive weights.}
    	Visual and LiDAR features are extracted and weighted so that distinctive modality contributes more in the combined descriptor.
    	The query and the true match have similar adaptive weights, both of which emphasize less on image features for their poor quality for recognition, while the wrong match has lower weight in LiDAR for the featureless corridor.
    	Darker colors represent relatively larger values.
    	}
    	\label{fig:idea}
    \end{figure}

    Aimed at different hardware settings and application scenarios, researchers have utilized diverse sensors for place recognition, among which to be used frequently are cameras and LiDARs. Vision-based methods \cite{VPR:cnnSeqSLAM, VPR:R-MAC, VPR:fine-tune-GeM, PR:netvlad} extract features from images and usually benefit from the abundant information that images encode. Their weaknesses, however, are also obvious that they are prone to light condition, season change and weather, especially in outdoor environments. In contrast, LiDARs are more robust to the variation of light and time \cite{PR:pointnetvlad}, and several LiDAR-based methods \cite{Feature3D:M2DP, PR:pointnetvlad, PR:LPDNet, yin2021fusionvlad} have been developed for long-term localization. Despite the precise geometric structure information from point clouds, they may encounter failure in some degenerate places like corridors and tunnels. For most applications, robots seem to have a single and specific working scenario so that a proper type of sensors can be chosen. But that is not always the case for some tasks, e.g. food or package delivery robots that work in both indoor and outdoor environments. Considering the merits of these two sensors, recent studies start to focus on the combination of 2D and 3D data in a way that concatenates \cite{VL-PR:augmenting} or fuses image and point cloud features with fully-connected (FC) layers \cite{VL-PR:oneshot}. Although the cooperation with another sensor effectively brings improvement in performance, one issue remains unsolved -- i.e., the two modalities are regarded as equally important at every time and every place.
    
    To handle this issue, it is natural to adjust the importance based on salient regions of the images and point clouds. One way enabling networks to learn where to look is to introduce attention. Unlike convolution operation, the attention mechanism focuses on long range interaction and is able to weight about the important or unnecessary features \cite{bello2019attention, fu2019dual, woo2018cbam}. Several works have applied it to the task of place recognition, but they only concentrate on the improvement on feature representation, either in a vision-based \cite{khaliq2019camal, chen2018learning} or LiDAR-based way \cite{PR:PCAN, xia2021SOE-Net, barros2021AttDLNet}. These attention augmented methods can, to some extend, strengthen the ability of feature extraction and make the recognition system more robust to appearance changes. Recently Lu \textit{et al.} \cite{lu2020PIC-Net} combines the visual and the LiDAR branches that are respectively augmented with attention to exploit the complementary advantages of the two modalities. Still, the modality contribution issue is untouched, and feature from less distinctive modality in a certain environment may worsen the performance of the whole system.
    
    In this paper, we propose AdaFusion, a multi-modality fusion network that learns the compact feature representation of both images and point clouds and then adjusts their contribution in different environmental situation. As shown in Fig. \ref{fig:idea}, AdaFusion mainly contains two branches: a feature extraction branch and a weight generation branch. The feature extraction branch encodes images and point clouds into distinctive descriptors separately, while the weight generation branch leverages the attention mechanism to learn the importance of different modalities. Moreover, to fully exploit the information in 2D and 3D data, these two branches are designed to work cooperatively. Multi-scale spatial and channel attention is first computed from different layers of the feature extraction branch and then fused together to produce adaptive weights. After that weighted features are concatenated to form a global descriptor. The contributions of our work are threefold:
    \begin{itemize}
    	\item We propose AdaFusion, a visual-LiDAR based place recognition network with adaptive weights. The weights serve as dynamic adjustment to the contribution of the two modalities in different environments, making our method outperforms current state-of-the-art works.
    	\item We design a weight generation branch that evaluates the input image and point cloud, and outputs the adaptive weights. The branch cooperates with the feature extraction branch and can be trained in an end-to-end manner.
        \item Experiments on two public datasets are conducted to evaluate the performance of our method under varying environmental situation.
	\end{itemize}
    Note that the adaptive weights improve the robustness of the system, relieving part of the stress to design more complicated feature extraction structures. We show in the experiments that simple feature extraction branch and low-dimension descriptor are still able to achieve acceptable recognition performance with the adaptive weights, making our method suitable to be deployed in low-cost mobile robot systems.

\section{Related Work}
    Generally place recognition methods encode sensor data into global descriptors which can be used for the retrieval of similar places with proper distance metric in the feature space. The encoding ways vary in terms of different sensors, hence we will have a brief review on the most frequently used types, i.e. the vision-based, the LiDAR-based and the visual-LiDAR fusion place recognition. 
    
    \textbf{Vision-based place recognition.} Visual place recognition has been studied for over two decades. Traditionally, handcrafted local feature descriptors \cite{PF:SIFT, bay2006surf} are utilized to capture salient information of the image. Later, with the booming of deep learning, researchers start to replace the handcrafted ones with networks \cite{VPR:cnnSeqSLAM, VPR:R-MAC, VPR:fine-tune-GeM} for their strong power in extracting features. The occurrence of NetVLAD \cite{PR:netvlad}, a trainable framework combining CNNs and VLAD, has inspired work such as Patch-NetVLAD \cite{VPR:PatchNetVLAD} to implement both local and global descriptors in an end-to-end manner. Regarding the learning-based methods, some work has incorporated the attention mechanism \cite{bello2019attention, fu2019dual, woo2018cbam} to enhance the resistance to visual appearance changes, since the context-aware attention distinguishes the important and unimportant regions of the image. For instance, Chen \textit{et al.} \cite{chen2018learning} proposed a context flexible attention model and fuse multi-scale attention maps into a final mask. CAMAL \cite{khaliq2019camal} used similar multi-scaling technique but with proposed regional attention.

    \textbf{LiDAR-based place recognition.} LiDARs can capture geometric structural information of the environment as point clouds that are sparse and unordered. Their irregular format, however, is hard to be applied to CNNs. One of the solution is first discretizing 3D space into voxel grids and then applying 3D convolution \cite{voxnet, shapenet}. Another way is brought by PointNet \cite{PointNet} that can directly take point clouds as input. Since then a few works have been proposed based on PointNet. For example, PointNetVLAD \cite{PR:pointnetvlad} combines PointNet and NetVLAD \cite{PR:netvlad} to enable an end-to-end training and extraction for global descriptor from 3D point clouds. Others like LPD-Net \cite{PR:LPDNet} and FusionVLAD \cite{yin2021fusionvlad} further deal with the feature aggregation and the viewpoint difference problems, respectively. Similar to vision-based methods, some researchers have applied the attention mechanism to networks for better concentration on important features, such as PCAN \cite{PR:PCAN}, SOE-Net \cite{xia2021SOE-Net} and AttDLNet \cite{barros2021AttDLNet}.
    
    \textbf{Visual-LiDAR fusion place recognition.} The idea of taking advantage of different modalities has recently caught researchers' eyes. Since cameras and LiDARs capture different information from the environment, the combination of the two sensors indeed has strengths over other single-modality methods, as reported in \cite{VL-PR:oneshot, VL-PR:augmenting, VL-PR:minkloc++, lu2020PIC-Net}. Ratz \textit{et al.} \cite{VL-PR:oneshot} project segments of point clouds onto the image and use 2D and 3D convolution to extract features. Features of different modalities are then fused by FC layers. Oertel \textit{et al.} \cite{VL-PR:augmenting} and Komorowski \textit{et al.} (MinkLoc++) \cite{VL-PR:minkloc++} adopt similar pipeline but without the projection and FC layers. Instead, the image and the point cloud features are concatenated directly. PIC-Net \cite{lu2020PIC-Net} further enhances the representation ability of features by integrating attention modules to the two feature extraction branches respectively. The function of attention is limited in each branch and the two modalities contribute equally, even when the environment is unfriendly for one of the sensors, say foggy days for cameras or corridors for LiDARs. In such cases advantages brought by multiple modalities may be lowered by the less informative sensor.

    \begin{figure*}[!t]
        \centering
        \includegraphics[width=0.9\linewidth]{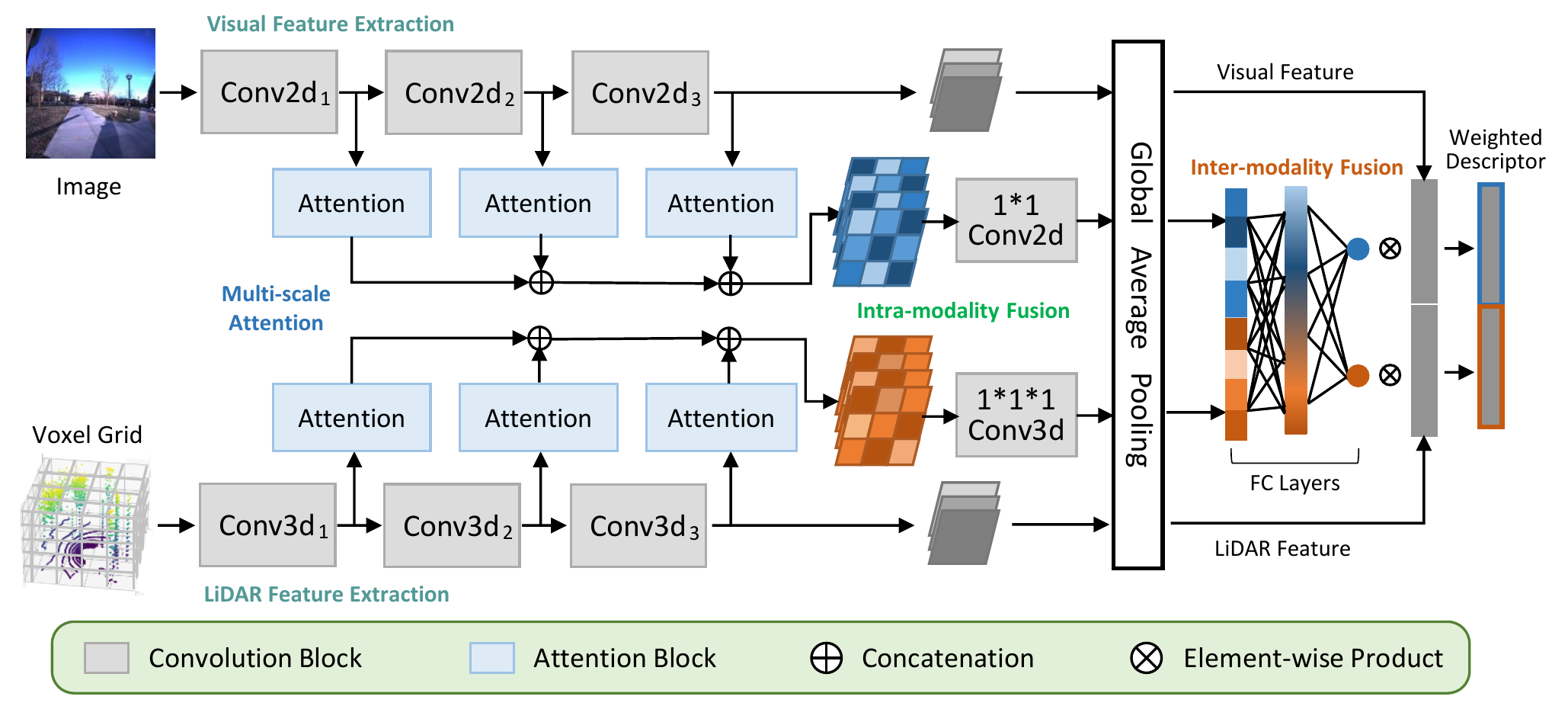}
        \put(-422,151){\footnotesize $\boldsymbol{I}$}
        \put(-416,78){\footnotesize $\boldsymbol{G}$}
        \put(-157,186){\footnotesize $\boldsymbol{M}_{\mathrm{I}}$}
        \put(-159,44){\footnotesize $\boldsymbol{M}_{\mathrm{P}}$}
        \put(-58,187){\footnotesize $\boldsymbol{f}_\mathrm{I}$}
        \put(-58,44){\footnotesize $\boldsymbol{f}_\mathrm{P}$}
        \put(-65,142){\footnotesize $\alpha_\mathrm{I}$}
        \put(-65,85){\footnotesize $\alpha_\mathrm{P}$}
        \put(-24,176){\footnotesize $\boldsymbol{f}'$}
        \caption{\textbf{The network structure of AdaFusion.}
        Our network is mainly composed of two branches.
        The feature extraction branch separately deals with images and point clouds to extract visual and LiDAR features with convolution.
        The weight generation branch firstly performs intra-modality fusion of the multi-scale attention, and then learns the adaptive weights with inter-modality fusion.}
        \label{fig:network_structure}
    \end{figure*}
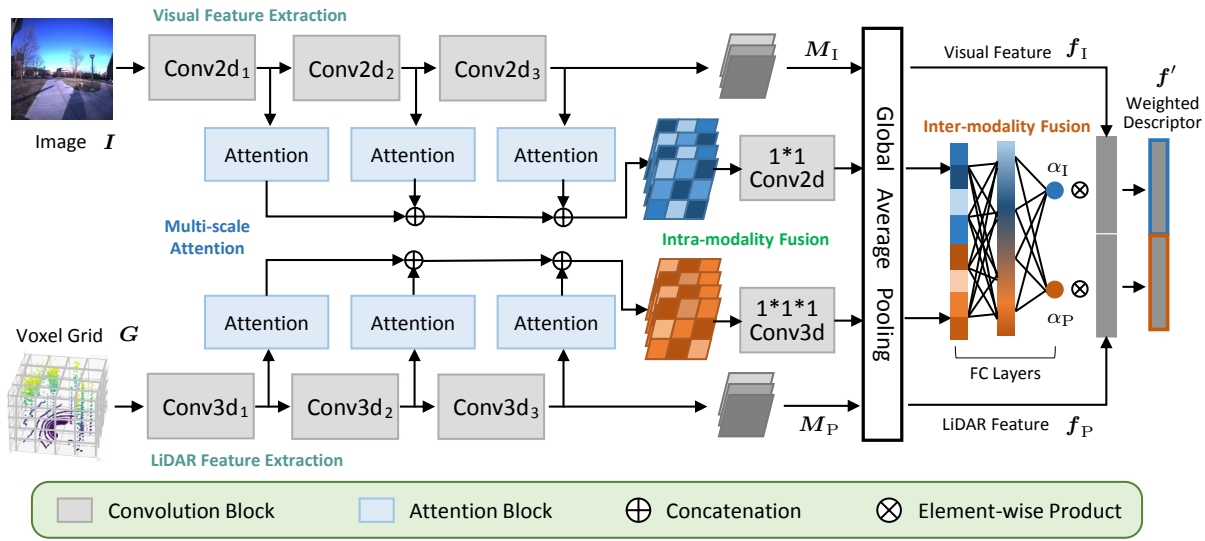

\section{Proposed Method}
    In this section, we show how AdaFusion leverages the weighted global descriptor to perform place recognition in varying environments. A brief illustration of the whole approach is shown in Fig. \ref{fig:idea}. Instead of simply extracting features from images and point clouds, the network also adjust the importance of these two modalities. After that, nearest neighbor (NN) search is used to retrieve the most similar places in the database with the weighted descriptor. Please note that the adaptive weights in Fig. \ref{fig:idea} have been divided by their average values, resulting in the percentage form. Further explanation can be found at Sec. \ref{subsec:PR_results}.

\subsection{Network Overview}
    There are two main aspects in AdaFusion, each of which is handled by a specific branch. The first one is feature extraction. Given a frame of image and point cloud, this branch utilizes 2D and 3D convolution to extract local visual and LiDAR features respectively in a parallel way, as shown in Fig. \ref{fig:network_structure}. Then these local features are aggregated to be global features using global average pooling (GAP). Although some techniques such as NetVLAD core \cite{PR:netvlad} can enhance the ability of feature extraction, we leave them out for two reasons. Some researchers \cite{VL-PR:oneshot, VL-PR:augmenting} have shown that simple convolution structures can also achieve high performance, which gets verified in our experiments. Moreover, the adaptive weights are able to adjust the importance and contribution of visual and LiDAR features so that the performance gets improved. Unlike methods that solely rely on feature extraction to output global descriptors, the existence of adaptive weights relieves part of the pressure to design more powerful but complicated feature extraction branches.

    Another aspect is the adaptive weights handled by the weight generation branch. As in Fig. \ref{fig:network_structure}, this branch contains three parts, namely multi-scale attention, intra-modality fusion and inter-modality fusion. The multi-scale attention is learned from different layers of the network. It has been reported that lower layers of the network often focus on fine details while those higher ones correspond to semantic meaning \cite{VPR:cnnSeqSLAM}. This property benefits the later fusion process because information at different scales is considered. Since it may be difficult for the network to directly learn the importance of modalities from their multi-scale attention, we design a two-stage fusion approach. Attention is first merged within the same type of sensor, summarizing information of the specific modality. After dimension reduction by GAP, FC layers are utilized to perform inter-modality fusion and output the adaptive weights. Given the weights and features, it is easy to get the weighted global descriptor with element-wise product.

\subsection{Feature Extraction}
    In this part we show how to obtain features from images and point clouds. We adopt the PyTorch tradition to denote high-dimensional array and omit the batch size for better illustration. For instance, 3D data is represented as $C\times H\times W$ and 4D data as $C\times D\times H\times W$, where $C,D,H$ and $W$ means \emph{Channel, Depth, Height} and \emph{Width} respectively.
    
    \textbf{Data preparation.} 
    Each frame $\boldsymbol{X}=(\boldsymbol{I},\boldsymbol{P})$ contains an RGB image $\boldsymbol{I} \in\mathbb{R}^{3\times H\times W}$ and point cloud $\boldsymbol{P} \in\mathbb{R}^{3\times N}$. For images we normalize them so that $\boldsymbol{I}$ is in range $[-1.0,1.0]$. However, the original data structure of $\boldsymbol{P}$ cannot be handled by convolutional operation. Instead, voxel grid $\boldsymbol{G} \in\mathbb{R}^{1\times X\times Y\times Z}$ with binary occupancy is utilized as the input, i.e. a voxel equals 1 if there is any point within it and 0 otherwise. Although other occupancy representations like point count, soft occupancy \cite{VL-PR:augmenting} and TDF \cite{FEATURE:3dmatch} are also available, the binary one has been shown to be more robust and has better recognition performance \cite{VL-PR:augmenting}.

\begin{figure}[t]
        \centering
        \includegraphics[width=0.9\linewidth]{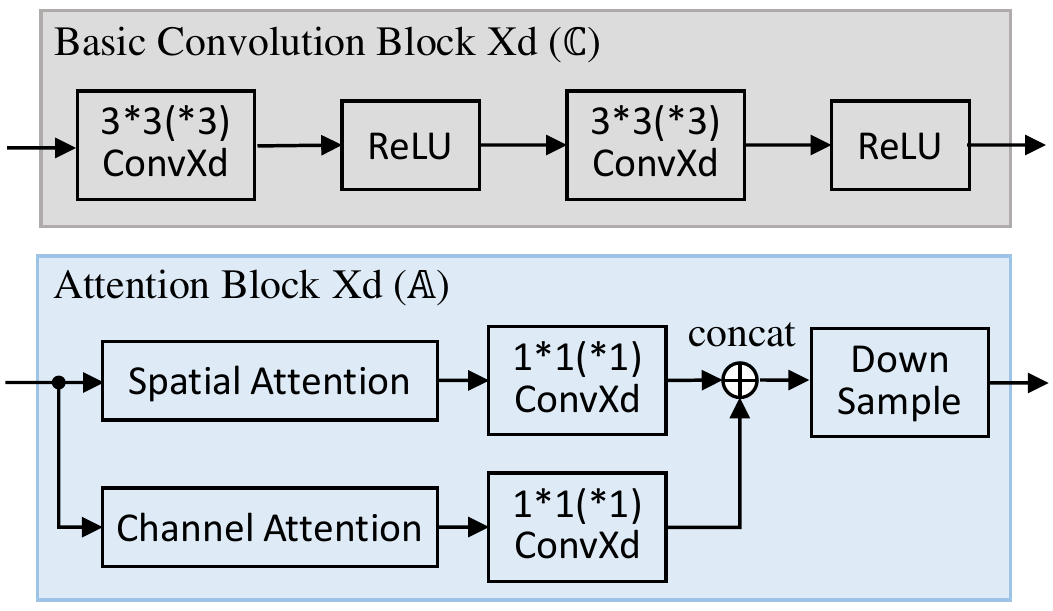}
        \caption{\textbf{Details of blocks.}
        The basic convolution block ($\mathbb{C}$) and the attention block ($\mathbb{A}$) are used to build the network, where $X$ can be either 2 or 3 depending on the data structure.
        The numbers in ConvXd mean the kernel size, and for all convolution the stride is 1.
        Concatenation is along the channel dimension of the feature map.}
        \label{fig:basic_blocks}
    \end{figure}

    \textbf{Structure and blocks.} 
    Both the backbones of visual and LiDAR feature extraction are divided into three convolution blocks $\mathrm{ConvXd}_i, X\in\{2,3\}, \; i=1,2,3$ which are further constituted by basic convolution blocks (denoted as $\mathbb{C}$). The structure of $\mathbb{C}$ is shown in Fig. \ref{fig:basic_blocks}, where convolution with kernel=3, stride=1 and padding=1 followed by ReLU activation function are repeated twice. Details of convolution blocks $\mathrm{ConvXd}_i$ are summarized in Fig. \ref{fig:comp_of_conv}. Note that we add a batch normalization (BN) layer \cite{batchnorm} for the last blocks $\mathrm{ConvXd}_3$ to reduce the internal covariate shift before getting the local visual feature map $\boldsymbol{M}_{\mathrm{I}} = (m^{\mathrm{I}}_{c,h,w}) \in \mathbb{R}^{C_1\times H_1\times W_1}$ and the local LiDAR feature map $\boldsymbol{M}_{\mathrm{P}} = (m^{\mathrm{P}}_{c,x,y,z}) \in \mathbb{R}^{C_1\times X_1\times Y_1\times Z_1}$. The use of BN in our structure improves the ability of feature extraction a lot. We observe that when directly using single visual or LiDAR feature as global descriptor for searching nearest neighbors, about 10\% improvement can be achieved with the BN layer compared to those without.
    
    \textbf{Global features.} 
    To produce the global visual feature $\boldsymbol{f}_\mathrm{I} = [f_\mathrm{I}^1, f_\mathrm{I}^2, \cdots, f_\mathrm{I}^{C_1}]^\top$ and the global LiDAR feature $\boldsymbol{f}_\mathrm{P} = [f_\mathrm{P}^1, f_\mathrm{P}^2, \cdots, f_\mathrm{P}^{C_1}]^\top$, we apply global average pooling (GAP) \cite{VPR:fine-tune-GeM} to $\boldsymbol{M}_{\mathrm{I}}$ and $\boldsymbol{M}_{\mathrm{P}}$, i.e.
    \begin{align}
        \label{eq:GAP_2D}  f_\mathrm{I}^c &= \frac{1}{H_1 W_1} 
            \sum_{h=1}^{H_1}\sum_{w=1}^{W_1}{m^{\mathrm{I}}_{c,h,w}}, &c=1,\cdots,C_1 \\
        \label{eq:GAP_3D}  f_\mathrm{P}^c &= \frac{1}{X_1 Y_1 Z_1}
            \sum_{x=1}^{X_1}\sum_{y=1}^{Y_1}\sum_{z=1}^{Z_1}{m^{\mathrm{P}}_{c,x,y,z}}, &c=1,\cdots,C_1
    \end{align}
    In fact, GAP is a special case of generalized-mean (GeM) pooling \cite{VPR:fine-tune-GeM}. Compared to FC layers, the global pooling methods have less parameters and are more robust to resolution changes of the input data.
    
    \begin{figure}[t]
        \centering
        \begin{tabular}{cc||cc}  
    		\hline
    		Block & Detail & Block & Detail \\
    		\hline
    		$\mathrm{Conv2d}_1$ & $\mathbb{C}_{64} \mathbb{P}_\mathrm{M} \mathbb{C}_{64}
    		    \mathbb{P}_\mathrm{M}$ & $\mathrm{Conv3d}_1$ & $\mathbb{C}_{32} \mathbb{P}_\mathrm{A}$ \\
    		$\mathrm{Conv2d}_2$ & $\mathbb{C}_{64} \mathbb{C}_{128} \mathbb{P}_\mathrm{M}$ &
    		    $\mathrm{Conv3d}_2$ & $\mathbb{C}_{64} \mathbb{C}_{64} \mathbb{P}_\mathrm{A}$ \\
    	    $\mathrm{Conv2d}_3$ & $\mathbb{C}_{128} \mathbb{C}_{128} \mathbb{B} \mathbb{P}_\mathrm{M}$ &
    		    $\mathrm{Conv3d}_3$ & $\mathbb{C}_{128} \mathbb{B} \mathbb{P}_\mathrm{A}$ \\
    		\hline
	    \end{tabular}
        \caption{\textbf{The composition of feature extraction structures.}
        We denote basic convolution block ($\mathbb{C}$) with $k$ output channels as $\mathbb{C}_k$, max pooling as $\mathbb{P}_\mathrm{M}$, average pooling as $\mathbb{P}_\mathrm{A}$ and batch normalization as $\mathbb{B}$.
        For all pooling operations, both their kernel and stride equal 2.}
        \label{fig:comp_of_conv}
    \end{figure}

\subsection{Adaptive Weights}
    We propose a weight generation branch that utilizes a two-stage fusion strategy to combine the attention information of images and point clouds. Slightly different from other attention augmented place recognition methods, the attention mechanism in our network does not serve as salient region masks. Instead, weights $\boldsymbol{\alpha}= [\alpha_\mathrm{I}, \alpha_\mathrm{P}]^\top$ are produced to change the contribution of visual and the LiDAR features.
    
    \textbf{Multi-scale attention.}
    Like \cite{chen2018learning}, multi-scale attention is used to fully exploit the information in different network layers. As shown in Fig. \ref{fig:network_structure} and Fig. \ref{fig:basic_blocks}, for each modality we compute the spatial attention and the channel attention illustrated in \cite{fu2019dual} from the feature extraction branch. However, in \cite{fu2019dual} and other related work \cite{barros2021AttDLNet, chen2018learning, woo2018cbam, bello2019attention}, the attention only aims at 2D images. We hence extend it to 3D voxel grid form. Given the query map $\boldsymbol{Q}$, the key map $\boldsymbol{K}$ and the value map $\boldsymbol{V}$ of shape $C_2\times X_2\times Y_2\times Z_2$, we first reshape them to shape $C_2\times N$, where $N=X_2\times Y_2\times Z_2$. Then the spatial attention map $\boldsymbol{S}_s \in \mathbb{R}^{N\times N}$ and the channel attention map $\boldsymbol{S}_c \in \mathbb{R}^{C_2\times C_2}$ are derived as 
    \begin{equation}
        \boldsymbol{S}_s = \texttt{Softmax}(\boldsymbol{K}^\top \boldsymbol{Q}), \quad
        \boldsymbol{S}_c = \texttt{Softmax}(\boldsymbol{Q} \boldsymbol{K}^\top),
    \end{equation}
    where the softmax operation is performed along the row (second dimension) of the matrix. Finally the spatial attention and the channel attention can be obtained by reshaping $\boldsymbol{V} \boldsymbol{S}_s^\top$ and $\boldsymbol{S}_c \boldsymbol{V}$ back to shape $C_2\times X_2\times Y_2\times Z_2$.
    
    The original spatial and channel attention in \cite{fu2019dual} have the same dimension as the input feature map, which may have large channel numbers when computing from higher layers of the network. To reduce computational cost and only keep useful information, we apply a kernel 1 convolution to linearly fuse the attention, as shown in Fig. \ref{fig:basic_blocks}. Before outputting from the attention block ($\mathbb{A}$), down sample with nearest interpolation is performed to ensure all the results have the same size.
    
    \textbf{Intra-modality fusion.}
    Now the multi-scale attention of the same modality is concatenated along the channel dimension. As mentioned above, attention from different layers of the network concentrates on varying contents. To fuse and merge them, a convolutional layer whose kernel equals 1 is again applied to the attention of the same modality. This operation acts as a transformation of the number of channels.
    
    \textbf{Inter-modality fusion.}
    One issue concerning fusing attention of different modalities is the inconsistency of the data structure. For example, in our case, the visual attention $\boldsymbol{A}_\mathrm{I}$ is of 3D shape ($C_3\times H_3\times W_3$) while the LiDAR attention $\boldsymbol{A}_\mathrm{P}$ is of 4D shape ($C_3\times X_3\times Y_3\times Z_3$). To handle this problem, we apply GAP to $\boldsymbol{A}_\mathrm{I}$ and $\boldsymbol{A}_\mathrm{P}$ respectively, which yields two 1D vectors $\boldsymbol{a}_\mathrm{I} \in \mathbb{R}^{C_3}$ and $\boldsymbol{a}_\mathrm{P} \in \mathbb{R}^{C_3}$ that have the same shape and dimension. The length of these vectors can be easily controlled in the intra-modality fusion stage, where $C_3=128$ is used in our network.
    
    Since $\boldsymbol{a}_\mathrm{I}$ and $\boldsymbol{a}_\mathrm{P}$ are now of the same dimension, a FC layer is applied so that $\boldsymbol{\alpha} = {\normalsize \texttt{FC}} \left([\boldsymbol{a}_\mathrm{I}^\top, \boldsymbol{a}_\mathrm{P}^\top]^\top \right),$ where $\boldsymbol{\alpha}=[\alpha_\mathrm{I}, \alpha_\mathrm{P}]^\top$ are the adaptive weights for visual and LiDAR features. The number of nodes from the input to the output layer is $[256,64,32,2]$, with two hidden layers applied. All activation functions are ReLU, except that Sigmoid is used for the last one so as to make $\alpha_\mathrm{I}, \alpha_\mathrm{P} \in [0,1]$. Finally the weighted global descriptor is derived as 
    \begin{equation}
        \boldsymbol{f}' = [\alpha_\mathrm{I} \boldsymbol{f}_\mathrm{I}^\top, \;
            \alpha_\mathrm{P} \boldsymbol{f}_\mathrm{P}^\top ]^\top,
    \end{equation}
    which can then be used to perform NN search for retrievals.

\subsection{Implementation Details}
\label{subsec:impl_details}
    When receiving a frame $\boldsymbol{X}=(\boldsymbol{I}, \boldsymbol{P})$, some pre-processing is made before feeding it to the network. We first remove ground points from the point cloud $\boldsymbol{P}$ and then convert it into voxel grid $\boldsymbol{G}$ with a resolution of $72\times 72\times 48$ voxels along the $X,Y,Z$ dimension respectively. For image $\boldsymbol{I}$, regions containing less useful information, like part of the data collection car or the ground, are cropped and removed. The remain is resized to resolution $300\times 400$ for $H\times W$. Data augmentation is also considered to enlarge the training set and to avoid overfitting. We randomly change the brightness, contrast and saturation of the image, and jitter the points in $\boldsymbol{P}$ before converting it to the voxel grid.
    
    Euclidean distance metric $\mathrm{d}_2(\cdot)$ is adopted to determine the true match (a.k.a. true positive, TP) and the wrong match (a.k.a. true negative, TN) of a query frame. Two frames $\boldsymbol{X}_i$ and $\boldsymbol{X}_j$ are deemed to be true match if their ground truth positions $\boldsymbol{x}_i$ and $\boldsymbol{x}_j$ are within a certain range, i.e. $\mathrm{d}_2( \boldsymbol{x}_i, \boldsymbol{x}_j) \le d_{\mathrm{pos}}$. Similarly for the wrong match, it has $\mathrm{d}_2( \boldsymbol{x}_i, \boldsymbol{x}_j) \ge d_{\mathrm{neg}}$. When training, we set $d_{\mathrm{pos}}=10$ m and $d_{\mathrm{neg}}=50$ m, while $d_{\mathrm{pos}} = d_{\mathrm{neg}} = 20$ m during testing.
    
    \textbf{Loss function.}
    In the field of place recognition, metric learning techniques are applied since there are no specific classes. Some popular loss functions, such as contrastive loss \cite{contrastive_loss}, triplet loss \cite{triplet_loss} and lazy triplet loss \cite{PR:pointnetvlad}, have been widely used in this task. The basic idea is that descriptors between true match should be pulled nearer, while those between wrong match should be pushed away. Here we adopt the pairwise margin-based loss with $L_1$ distance metric \cite{VL-PR:augmenting}:
    \begin{equation}
        \mathcal{L}(\boldsymbol{f}'_i, \boldsymbol{f}'_j, y) = \left\{
        \begin{aligned}
            \left[\mathrm{d}_1(\boldsymbol{f}'_i, \boldsymbol{f}'_j) - (m-a)\right]_+, \ y&= 1, \\
            \left[(m+a) - \mathrm{d}_1(\boldsymbol{f}'_i, \boldsymbol{f}'_j)\right]_+, \ y&=-1,
        \end{aligned} 
        \right.
    \end{equation}
    where $[x]_+=\max(x,0)$ is the hinge loss. The label $y=1$ for true match and $y=-1$ otherwise. It is shown in \cite{VL-PR:augmenting} that the pairwise margin-based loss can achieve good performance and converge faster.

    \textbf{Training configuration.}
    Our network is implemented in PyTorch and trained with a single Nvidia RTX 3090 GPU and 64 GB RAM. For each case, we train the network for 50 epochs and evaluate its performance every 2000 batches. The Adam optimizer is used with the initial learning rate set to $8 \times 10^{-4}$. It will be reduced by a factor of 0.9 if no improvement is achieved in the following 20 evaluations.

\section{Experiments}
    This section presents experimental results and analyses of AdaFusion. We compare it with state-of-the-art place recognition methods \cite{PR:netvlad, PR:pointnetvlad, VL-PR:augmenting, lu2020PIC-Net, VL-PR:minkloc++, VL-PR:CORAL} and show the advantages of our adaptive weights. All of the results reported about our method is based on the structure shown in Fig. \ref{fig:network_structure}. However, it should be noted that the performance of our network can be further improved by adding attention in the feature extraction branch. It is easy to implement and will not increase much computational cost owing to the reuse of attention blocks, but we leave it out for simplicity and better illustration of the adaptive weights.

\subsection{Datasets and Evaluation}
    \begin{table}[t]
    \centering
    \caption{Sequences used in the experiment and the number of frames, pairs and testing queries.}
    \begin{tabular}{c|cc|l}
    \hline
            & \multicolumn{2}{c|}{Sequences}            & Numbers$^{\rm 1}$ \\ \hline
    \multirow{5}{*}{\rotatebox{90}{RobotCar}}
            & 2014-07-14-14-49-50 & 2014-11-18-13-20-12 & \ding{172}: 5155 \\
            & 2014-12-02-15-30-08 & 2014-12-09-13-21-02 & \ding{173}: 2394 \\
            & 2014-12-12-10-45-15 & 2015-02-03-08-45-10 & \ding{174}: 65k  \\
            & 2015-02-13-09-16-26 & 2015-03-10-14-18-10 & \ding{175}: 14M  \\
            & 2015-05-19-14-06-38 & 2015-08-13-16-02-58 & \ding{176}: 21k  \\ \hline
    \multirow{5}{*}{\rotatebox{90}{NCLT}}
            & 2012-01-08          & 2012-01-22          & \ding{172}: 4356 \\
            & 2012-02-12          & 2012-02-18          & \ding{173}: 1111 \\
            & 2012-03-31          & 2012-05-26          & \ding{174}: 44k  \\
            & 2012-08-04          & 2012-10-28          & \ding{175}: 9M   \\
            & 2012-11-04          & 2012-12-01          & \ding{176}: 9.9k \\ \hline
    \end{tabular} \\
    \raggedright \vspace{1pt} \hspace{10pt}
    {$^{\rm 1}$ \scriptsize \ding{172}: frames in training set, \ding{173}: frames in testing set, \ding{174}: true match pairs, \\ \hspace{17pt} \ding{175}: wrong match pairs, \ding{176}: testing queries.}
    \label{table:dataset_summary}
    \end{table}
    
    \begin{figure}[t]
        \includegraphics[width=0.95\linewidth]{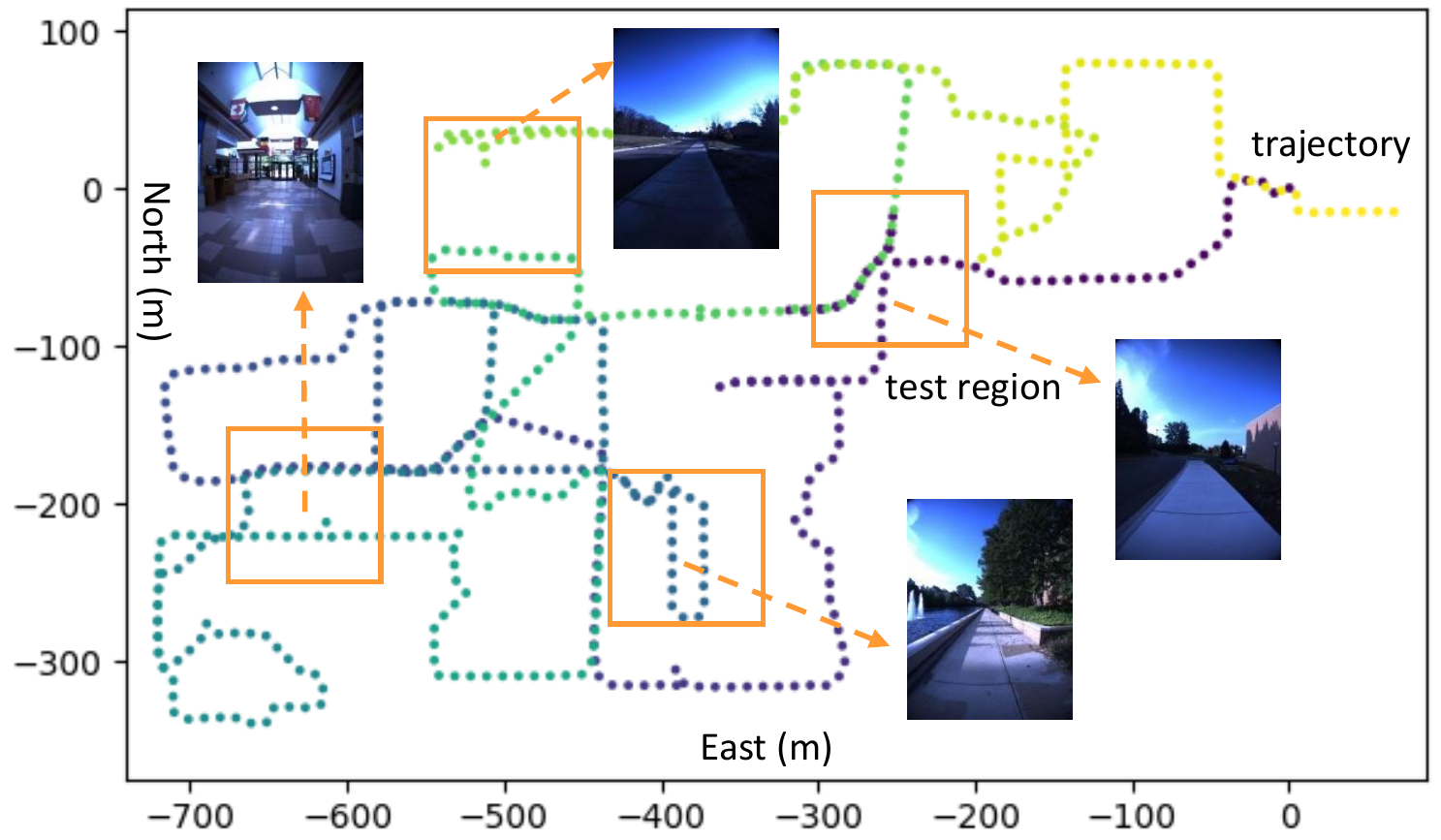}
        \caption{\textbf{The trajectory and the training \& testing set of the NCLT dataset.}
        Dots in the trajectory represent selected frames, and orange boxes with size 100 m $\times $ 100 m are test regions forming the testing set. The training set is comprised of the remaining disjoint region.}
        \label{fig:test_regions}
    \end{figure}

    The proposed method is tested on two public datasets, namely Oxford RobotCar \cite{DATASET:RobotCar} and NCLT \cite{DATASET:NCLT}.
    
    \textbf{Oxford RobotCar Dataset.}
    This dataset was repeatedly collected in the central Oxford, UK twice a week for more than one year, resulting in over 130 sequences. Due to the long time span, the sequences cover different outdoor conditions like season, weather and light changes. Although RGB images are provided, point clouds need to be generated by the accumulation of 2D laser points. For convenience we use the processed point clouds submaps from \cite{PR:pointnetvlad}, where each submap represents a 20 m trajectory of the car with 10 m overlap. Then we construct frames $\boldsymbol{X}$ by associating images to submaps using their timestamps.
    
    \textbf{NCLT Dataset.}
    Similar to the previous one, this dataset was collected in the University of Michigan's North Campus for 15 months and has 27 sequences. But what makes it different is that it contains both indoor and outdoor scenarios, and is more challenging in viewpoint difference. Since RGB images and 3D LiDAR point clouds are already provided in this dataset, we directly associate them as frames $\boldsymbol{X}$ according to the timestamps. To make configurations of these two datasets similar, we select frames from each sequence for every 10 m trajectory of the robot. 
    
    For both datasets, we respectively choose 10 sequences which cover different environmental conditions. A list of these sequences can be found in Table \ref{table:dataset_summary}, in which those of the RobotCar dataset are identical to \cite{VL-PR:augmenting} for fair comparison. Besides, to construct training and testing sets, sequences are divided into geographically disjoint regions. Some of the testing frames can be seen in Fig. \ref{fig:test_regions}, where we deliberately include both indoor and outdoor places. Using the method introduced in Sec. \ref{subsec:impl_details}, we exhaustively view all frames as the query and search other frames to make true match pairs and wrong match pairs. These pairs can then be used by the pairwise margin-based loss for training. When it comes to testing, every two sequences are chosen as query and database, giving rise to a total number of $\mathrm{C}_{10}^2=45$ combinations per dataset. Table \ref{table:dataset_summary} shows a summary of the number of frames, training pairs and testing queries.
    
    \textbf{Evaluation metric.}
    We report the \emph{recall@$N$} of all the evaluated methods. It represents the percentage that at least one true positive (i.e. true match) occurs in the top-$N$ retrievals of the query using KNN search. In particular, average recall@$1$ (AR@$1$) and average recall@$1\%$ (AR@$1\%$) of different methods are compared, where $N$ equals 1 and $1\%$ of the searching database respectively.
    
\subsection{Place Recognition Results}
\label{subsec:PR_results}
    \begin{table}[t]
    \centering
    \caption{Average Recall (\%) of the evaluated place recognition methods on the Oxford RobotCar dataset.}
    \begin{tabular}{lccc}
        \hline
        \multicolumn{1}{c}{Methods} & Modality$^{\rm 1}$ & AR@1     & AR@1\%        \\ \hline
        NetVLAD \cite{PR:netvlad}          & V     & 51.52          & 63.19         \\
        PointNetVLAD \cite{PR:pointnetvlad}& L     & 63.87          & 81.29         \\
        Ref. \cite{VL-PR:augmenting}       & V+L   & 98.00          & -             \\
        PIC-Net \cite{lu2020PIC-Net}       & V+L   & -              & 98.22         \\
        MinkLoc++ \cite{VL-PR:minkloc++}   & V+L   & 96.70          & 99.10         \\
        CORAL \cite{VL-PR:CORAL}           & V+L   & 88.93          & 96.13         \\
        AdaFusion (our)                    & V+L   & \textbf{98.18} & \textbf{99.21}\\ \hline
    \end{tabular} \\
    \raggedright \vspace{1pt} \hspace{25pt}
    {$^{\rm 1}$ \scriptsize V: Visual, L: LiDAR, V+L: Visual+LiDAR (multi-modality).}
    \label{table:results_of_methods}
    \end{table}
    
    Both single modality (Visual or LiDAR) and multi-modality place recognition methods are compared with our one. For NetVLAD \cite{PR:netvlad} and PointNetVLAD \cite{PR:pointnetvlad}, we run the evaluation by ourselves with the released source codes, while the results of \cite{VL-PR:augmenting}, PIC-Net \cite{lu2020PIC-Net}, MinkLoc++ \cite{VL-PR:minkloc++} and CORAL \cite{VL-PR:CORAL} are reported as in the original papers. Table \ref{table:results_of_methods} shows the recognition results on the Oxford RobotCar dataset. As we can see, all the compared multi-modality based methods outperform those based on single modality both in AR@1 and AR@1\%, which indicates the effectiveness of the fusion approach. Moreover, although the performance of state-of-the-art fusion based methods is already high, ours can still improve AR@1 by 0.18\% compared to \cite{VL-PR:augmenting} and AR@1\% by 0.11\% compared to MinkLoc++ \cite{VL-PR:minkloc++}, finally reaching 98.18\% and 99.21\% for AR@1 and AR@1\% respectively.

    \begin{figure}[t]
        \centering
        \includegraphics[width=\linewidth]{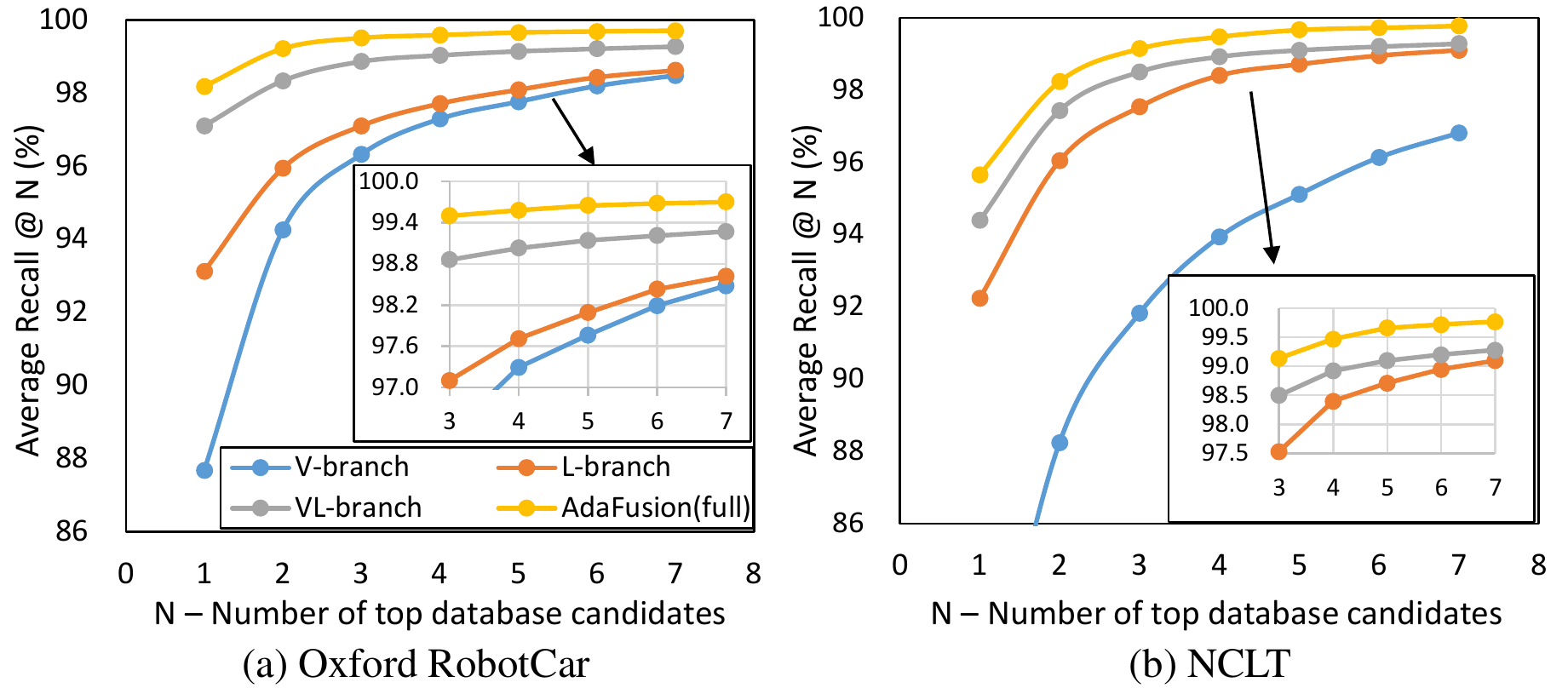}
        \caption{\textbf{AR@$N$ of the network.}
        We show the results tested on the (a) Oxford RobotCar and (b) NCLT datasets, where "V" and "L" are short for "visual" and "LiDAR" respectively. The blue, orange and gray curves illustrate the performance of our feature extraction branch, while the yellow one shows that of the whole AdaFusion.}
        \label{fig:recall_at_N}
    \end{figure}

    The performance of each branch in AdaFusion is tested on the Oxford RobotCar and the NCLT dataset and shown in Fig. \ref{fig:recall_at_N}. Since our feature extraction branch consists of two sub-branches, we separately train and evaluate the visual and the LiDAR feature extraction parts. The trend of AR@$N$ is similar in both datasets, where LiDAR feature (orange) is more distinctive than visual feature (blue), and the direct concatenation of the two modalities (gray) enhances the performance. When applying the adaptive weights (yellow), further improvement is achieved. The effect of adaptive weights is more significant with a small $N$, as the performance of feature extraction branch gets better and the difference becomes smaller when the number of candidates increases.

    \begin{figure*}[!t]
        \centering
        \includegraphics[width=0.9\linewidth]{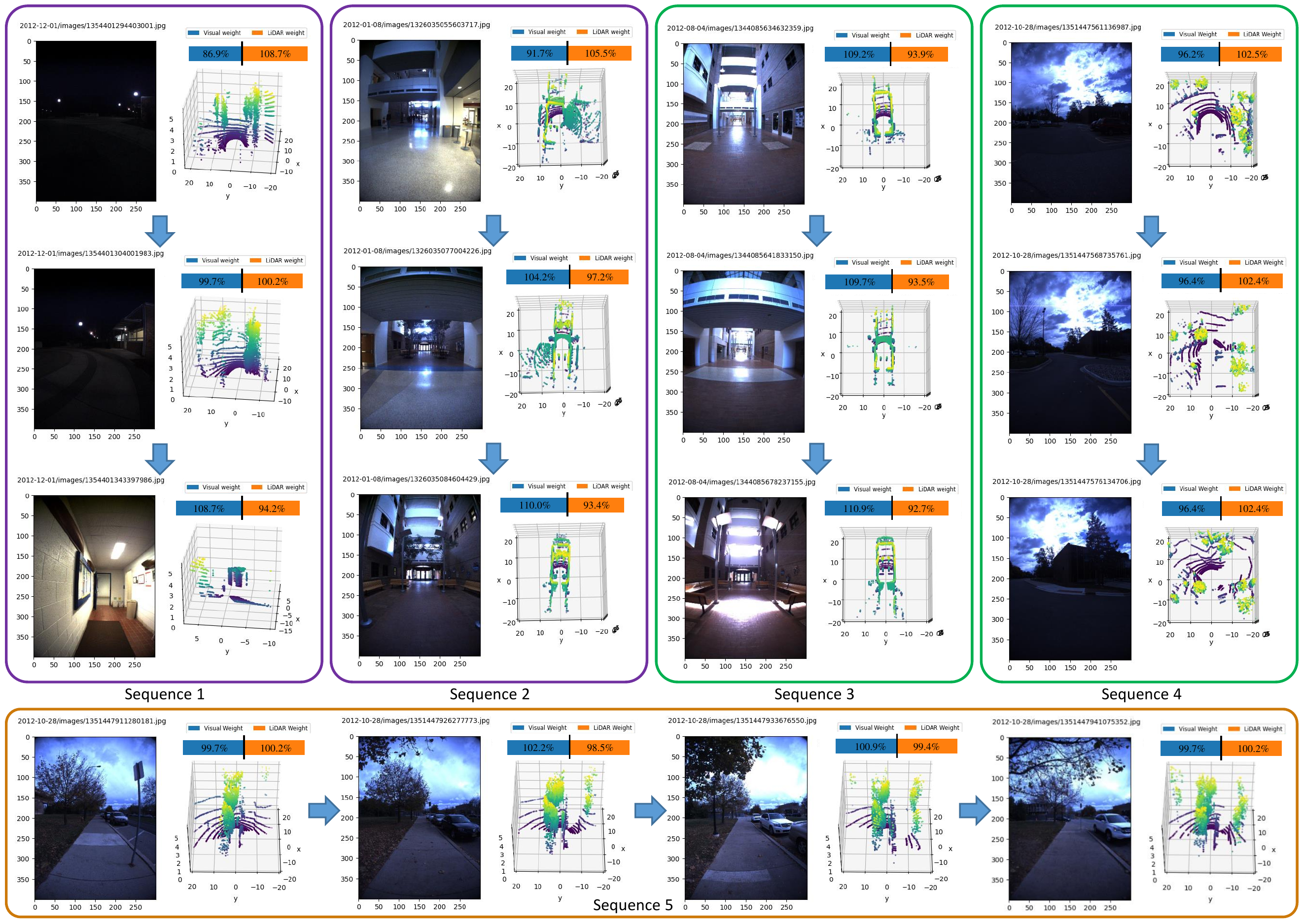}
        \caption{\textbf{Adaptive weights of different places.}
        There are consecutive frames in each sequence. The weights are converted to relative ratios compared to the average visual and LiDAR weights for better illustration. A value higher than $100\%$ means the corresponding modality contributes more in this place than the average contribution of all places.}
        \label{fig:weights_demo}
    \end{figure*}
    
    To show how AdaFusion benefits from the dynamic adjustment of the two modalities, we investigate the adaptive weights in a series of places. Fig. \ref{fig:weights_demo} presents five sequences, each of which contains consecutive frames while the robot moves. Sequence 1 and 2 show that the weights change as environment varies. When a robot enter the bright house from the dark outside, as in Seq. 1, the quality of images is getting better and thus the visual weight increases. Similarly, Seq. 2 suggests that the LiDAR weight would decrease when the robot moves into the corridor where only walls appear in the point cloud. Sequence 3 and 4 illustrate the dominating modality. For example, as shown in Seq. 3, indoor environment is usually unfriendly for LiDAR sensor because it is easy to get walls in the point clouds, so camera is likely to become the dominant sensor. On the contrary, Seq. 4 shows that point clouds are usually more distinctive in outdoor environments, so the LiDAR weight is relatively larger. However, there are some difficult cases like the one in Seq. 5, where places are unfriendly to both images and point clouds. We cannot easily distinguish them apart using whether visual or LiDAR information. As a consequence, the contribution of the two modalities is nearly equal in such case. Note that the percentages are neither the real weight values nor normalized ratios between the two modalities. In fact they are relative ratios compared to the average visual and LiDAR weights, i.e. $\bar{\alpha}_\mathrm{I} = 0.4$ and $\bar{\alpha}_\mathrm{P}=0.6$ for this dataset. Apart from better illustration, this representation can reflect the current contribution of a modality with respect to its average standard, because each query is searched through the whole database.
    
\subsection{Ablation Study}
    In order to explore the effects between the feature extraction branch and the weight generation branch, we evaluate the AR@1 of each component of our network with respect to different feature dimensions, as shown in Table \ref{table:feature_dimension}. Our baseline is based on feature dimension $d_f=256$ (D-256) where $d_{f_\mathrm{I}} = d_{f_\mathrm{P}} = d_f/2$. As we can see, the decrease in feature dimension indeed lowers the recognition performance of the feature extraction branch, especially for the V-branch. The large difference in performance between the visual and the LiDAR feature may hinder the cooperation of these two modalities in the fused descriptor. The adaptive weights, to some extend, balance the performance difference and further improve the recognition recall. The evidence is that we have observed that the average visual and LiDAR weights (i.e. $\bar{\alpha}_\mathrm{I}$ and $\bar{\alpha}_\mathrm{P}$) are changed from 0.40 and 0.60 when $d_f=256$ to 0.32 and 0.68 when $d_f=128$ for the NCLT dataset, which coincides with the decreased performance of the visual feature. It shows that the adaptive weights can improve the robustness of the system and help achieve acceptable results even with simple feature extraction branch and low-dimension descriptor.
    
    \begin{table}[t]
    \centering
    \caption{AR@1 (\%) for different feature dimensions, where D and Impr. are short for Dimension and Improvement, respectively.}
    \begin{tabular}{c|ccccc}
        \hline
            & Types     & D-128 & D-192 & D-256 & D-512 \\
        \hline
        \multirow{5}{*}{\rotatebox{90}{RobotCar}}
            & V-branch  & 83.57 & 86.87 & 87.68 & 88.91 \\
            & L-branch  & 89.47 & 91.45 & 93.12 & 93.27 \\
            & VL-branch & 96.35 & 96.59 & 97.10 & 97.32 \\
            & AdaFusion & 97.42 & 97.72 & 98.18 & 98.24  \\
            & Impr. & \textbf{1.07} & \textbf{1.13} & \textbf{1.08} & \textbf{0.92} \\
        \hline
        \multirow{5}{*}{\rotatebox{90}{NCLT}}
            & V-branch  & 69.48 & 73.79 & 79.34 & 80.23 \\
            & L-branch  & 90.75 & 91.07 & 92.23 & 93.54 \\
            & VL-branch & 92.61 & 93.35 & 94.39 & 95.81 \\
            & AdaFusion & 94.06 & 94.85 & 95.65 & 96.97 \\
            & Impr. & \textbf{1.45} & \textbf{1.50} & \textbf{1.26} & \textbf{1.16} \\
        \hline
    \end{tabular}
    \label{table:feature_dimension}
    \end{table}

\section{Conclusions}
This paper presents AdaFusion, an adaptive weighting visual-LiDAR fusion method for place recognition. To handle the issue that different modalities contribute equally even in environment unfriendly for one of them, we introduce the adaptive weights which dynamically adjust the importance of visual and LiDAR features in the global descriptor. The weights are learned from the weight generation branch which leverages multi-scale spatial and channel attention. Experiments on two public datasets show that our AdaFusion outperforms state-of-the-art fusion based methods. The presence of adaptive weights further improves the robustness of the system and makes it possible to obtain better results with less powerful but low-cost feature extraction structures.

\section{Acknowledgment}
The authors want to thank Beijing Novauto Technology Co., Ltd for generously providing GPUs for training.

\bibliographystyle{Bibliography/IEEEtranTIE}
\bibliography{RefAbrv}

\begin{IEEEbiography}[{\includegraphics[width=1in,height=1.25in,clip,keepaspectratio]{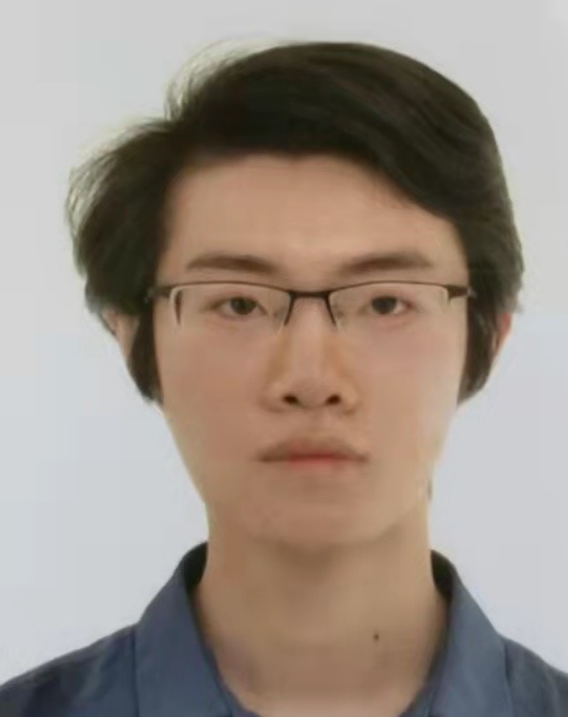}}]
    {Haowen Lai} received the B.E. degree in control science and engineering from Tongji University, Shanghai, China, in 2019. He is currently pursuing an M.S. degree from Tsinghua University, Beijing, China. His research field mainly covers 3D localization and perception, SLAM, Place Recognition, and their application in robotics. He is trying to apply computer vision and machine learning techniques to robotics so as to make robots more intelligent in understanding the environment and performing safe navigation.
\end{IEEEbiography}

\vspace{-1cm}
\begin{IEEEbiography}[{\includegraphics[width=1in,height=1.25in,clip,keepaspectratio]{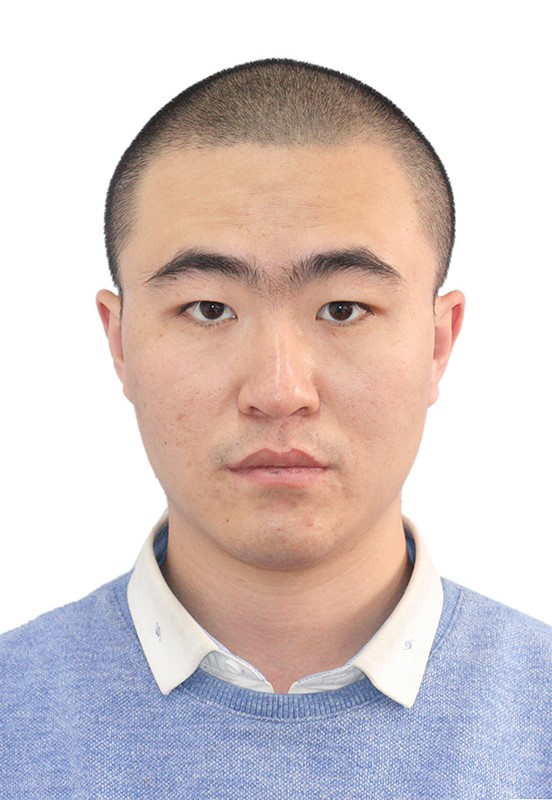}}]
    {Peng Yin} received the Bachelor degree from Harbin Institute of Technology, Harbin, China, in 2013, and the Ph.D. degree from the University of Chinese Academy of Sciences, Beijing, in 2018.
    He is a research Post-doctoral with the Department of the Robotics Institute, Carnegie Mellon University, Pittsburgh, USA.
    His research interests include LiDAR SLAM, Place Recognition, 3D Perception, and Reinforcement Learning. Dr. Yin has served as a Reviewer for several IEEE Conferences ICRA, IROS, ACC.
\end{IEEEbiography}

\vspace{-1cm}
\begin{IEEEbiography}[{\includegraphics[width=1in,height=1.25in,clip,keepaspectratio]{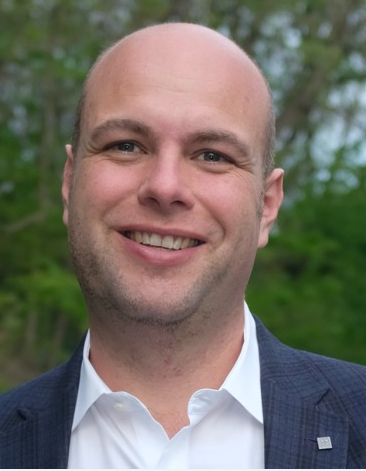}}]
    {Sebastian Scherer} received his B.S. in Computer Science, M.S. and Ph.D. in Robotics from CMU in 2004, 2007, and 2010. 
    
    Sebastian Scherer is an Associate Research Professor at the Robotics Institute at Carnegie Mellon University. His research focuses on enabling autonomy for unmanned rotorcraft to operate at low altitude in cluttered environments. He is a Siebel scholar and a recipient of multiple paper awards and nominations, including AIAA@Infotech 2010 and FSR 2013. His research has been covered by the national and internal press including IEEE Spectrum, the New Scientist, Wired, der Spiegel, and the WSJ. 
\end{IEEEbiography}

\end{document}